\documentclass[12pt,onecolumn]{article}
\usepackage{amssymb,amsmath,epsfig,graphics,graphicx,enumerate,float,subcaption,verbatim,xspace, color,url}
\usepackage{multirow}
\usepackage{epstopdf}

\newcommand{\nop}[1]{}

\newcommand{\eg}{{\sl e.g.}\xspace}

\begin{document}

\title{When Not to Classify: Detection of Reverse Engineering Attacks on DNN Image Classifiers}

\author{
Yujia Wang, David~J.~Miller, George Kesidis\thanks{This paper is dedicated to the memory of our dear friend Jan Larsen. This research is 
supported by AFOSR DDDAS
and Cisco Systems URP.}\\
~\\
School of EECS\\
The Pennsylvania State University\\
University Park, PA 16802\\
\{yjwang,djm25,gik2\}@psu.edu
}

\maketitle

\begin{abstract}
This paper addresses detection of a reverse engineering (RE) attack
targeting a deep neural network (DNN) image classifier; by
querying, RE's aim is to discover the classifier's
decision rule.
RE can enable
test-time evasion attacks, which require knowledge of the classifier.
Recently, we  proposed  a quite effective approach (ADA) to
detect test-time evasion attacks.
In this paper, we extend ADA to
detect RE attacks (ADA-RE). 
We demonstrate our method is successful in detecting ``stealthy" RE
attacks before they learn enough to launch effective test-time evasion attacks.
\end{abstract}
\section{Introduction}
Recently, there has been great interest in identifying vulnerabilities in machine learning (ML)) systems.
{\it Test-time evasion attacks} (TTEAs)
\cite{Papernot,Goodfellow,CW,Wagner17,ADA-arxiv17,MLSP18-ADA}
%Papernot = McDaniel16
add subtle perturbations to 
legitimate test-time samples\footnote{Perturbation 
approaches are related to boundary-finding algorithms
(neural network inversion) \cite{Hwang97}.}
to ``fool'' a classifier into making incorrect decisions relative to those of a human being.  
Related work has demonstrated the fragility of DNNs for some domains in the presence of modest data perturbations, e.g. 
changing the tempo in music genre classification [12]. %\cite{Larsen}.
TTEAs should be taken
seriously because they could
allow illegitimate access to a building, data, or a piece of machinery.  They could also lead e.g.
to a radiologist looking at ``doctored'' cancer biopsy images (that trick automated pre-screening systems).   
Test-time attacks require knowledge of the classifier under attack.
RE attacks \cite{Reiter16, Papernot_RE} involve
querying a classifier to {\it discover}
its decision rule.  Thus, one primary  
application of RE
is to {\it enable} TTEAs. 

Several recent RE attack works are \cite{Reiter16} and 
%\cite{PapernotMGJCS16} 
\cite{Papernot_RE}.
\cite{Reiter16} demonstrates that, with a {\it relatively} modest number of queries (perhaps $\sim$ ten thousand), using the classifier's answers on query examples as supervising ground truth labels, one can learn
a surrogate classifier on a given domain that closely mimics an unknown (black box) classifier. 
One weakness of \cite{Reiter16} is that it neither considers very large (feature
space) domains nor very large networks (DNNs) -- orders
of magnitude more queries may be needed to reverse-engineer a DNN on a large-scale domain.  However, a much more critical weakness stems from one of the greatest purported
advantages in \cite{Reiter16} -- the authors emphasize their RE does not require {\it any} actual
samples from the domain\footnote{For certain sensitive domains, or ones where obtaining real examples is expensive, the 
attacker may not have access to legitimate examples.}. Their queries
are {\it randomly} drawn, \eg uniformly, over the given feature space.  What was not recognized in \cite{Reiter16} is that
this random querying makes the attack {\it easily detectable} -- randomly
selected query patterns will typically look nothing like legitimate examples from any of the classes --
they are very likely to be extreme outliers, of all the classes.  Each such query is thus
{\it individually} highly suspicious -- thus, even ten, let alone ten thousand such queries
will be trivially anomaly-detected as jointly improbable under a null distribution (estimable from the training set defined over all the classes from the domain).  Even if the attacker employed bots, each of
which makes a small number of queries, each bot's random queries should be easily detected
as anomalous, likely associated with an RE 
attack. On the other hand,
\cite{Papernot_RE} propose an RE attack that does require some initial known data from the domain.  It uses this to create more
legitimate, ``stealthier'' queries, over a series of query stages, with the resulting labeled data used to train a substitute classifier used to launch a TTEA.

Recently, an approach was developed which achieves state-of-the-art results in detection of TTEAs,
Anomaly Detection of Attacks (ADA) \cite{ADA-arxiv17,MLSP18-ADA}.  Since this approach is an anomaly detector for the image domain of interest, it in principle
should also be applicable to detect query images that are not representative of real images from the domain.  However, since the querying in \cite{Papernot_RE} is stealthy (as it is based on perturbations of real images from the domain), it is not obvious their querying is detectable.  However, here
we extend ADA to indeed detect the RE querying from \cite{Papernot_RE}, and thus demonstrate the potential to prevent TTEAs even before
they are launched.

This paper is organized as follows. In Sec. 2, we describe
the reverse engineering attack of [11]. In Sec. 3, we give
background on ADA. In Sec. 4, we discuss our extension of
ADA for reverse engineering attacks. Experimental results
for DNN classifiers of images are given in Sec. 5. Finally,
conclusions are drawn in Sec. 6.
\section{RE Attack Given Domain Samples}\label{sec:back}
The RE procedure in
\cite{Papernot_RE} is summarized as follows.
First, the adversary collects a small set of representative labeled samples from the input domain as an initial training set $S_0$ and
uses this to train an initial substitute classifier.
Then, there is stagewise data collection and retraining, over a sequence of stages. In each, the adversary augments the current training set by querying the classifier with the stage's newly generated samples \cite{Papernot_RE}, i.e.,
\[S_{k+1} =\{\underline{x}+\lambda\cdot{\rm sgn}(\triangledown({\rm max}_cP^{(k)}_s[C=c|\underline{x}])):\underline{x}\in S_k\}\cup S_k\] 
where $k$ is the current stage index and $P^{(k)}_s[C=c | \underline{x}]$ is the current substitute class posterior model.  
The substitute classifier is then retrained using $S_{k+1}$.  Each successive stage crafts query samples closer to the classifier's true
boundary, which is helpful for RE learning but which also makes these samples less class-representative and thus more detectable.
Once a sufficiently accurate substitute classifier is learned, the adversary can launch a TTEA using one of the existing TTEA
attacks, e.g. \cite{Goodfellow,Papernot,CW}.  Here, one starts with an original image from the domain, from a source class $c_s$,
perturbs the image, using the substitute classifier's gradient information, to push across the decision boundary to a destination class, $c_d \neq c_s$.  The perturbed image is then submitted to the actual classifier as a TTEA instance.
\section{Detection of Test-Time Evasions (ADA)}\label{sec:ADA}
\subsection{Basic ADA}
ADA detection is grounded in the premise that an attack example in general will exhibit too much atypicality (evaluated on null distributions
estimated from the class training sets) w.r.t. $c_d$ and too little null atypicality w.r.t. $c_s$\footnote{This premise is plausible because the attacker tries to be stealthy -- to fool the classifier while not fooling a human being (or an anomaly detector).  In so doing, the perturbed image, while classified to $c_d$, still has to ``look'' like it comes from $c_s$.}. Given a test sample $\underline{x}$, basic ADA works as follows:
\begin{enumerate}
\item Determine the MAP (destination) class under the deep neural network: $c_d={\rm argmax}_cP[C=c|\underline{x}]$.
\item  Compute $\underline{z} = \underline{g}_k(\underline{x})$, the vector of outputs from the $k$-th layer of the DNN.
\item Estimate the source class $c_s$ based on the null model: $c_s = {\rm argmax}_{c \neq c_d} f_{\underline{Z}(k)|c} (\underline{z})$.
\item Form two probability vectors $P(k)$ and $Q(k)$ where $P(k) = \{ p_0 f_{\underline{Z}(k)|c_d} (\underline{z}),  p_0 f_{\underline{Z}(k)|c_s} (\underline{z}) \}$ and $Q(k) = \{ q_0 P[C=c_d | \underline{x}], q_0 P[C=c_s | \underline{x}] \}$. $p_0$ and $q_0$ are normalizers to make $P(k)$ and $Q(k)$ probability mass functions.
\item Report a detection if $D_{\rm KL}(P(k)||Q(k))>t$ where $D_{\rm KL}(\cdot||\cdot)$ is the Kullback Leibler (KL) Divergence.
\end{enumerate}
The KL distance will be large when $\underline{x}$ exhibits atypicality w.r.t. the null of $c_d$ and typicality w.r.t. the null of $c_s$.
\subsection{Ultimate ADA Method Development: L-AWA-maxKL}
The ultimate ADA method is based on the following extensions/improvements.

\paragraph{Maximizing KL distance over multiple layers:} Rather than measure KL distance at one layer, we can compute KL distance at different layers and detect based on the {\it maximum} KL distance over these layers.

\paragraph{Null modelling for Different Neuron Activations:} It was demonstrated that Gaussian mixture modelling is suitable for sigmoidal and linear layers, with log-Gaussian mixture modelling appropriate for RELU layers\cite{ADA-arxiv17,MLSP18-ADA}.

\paragraph{Exploiting source class uncertainty and class confusion:}
Basic ADA hard-estimates $c_s$. More information is preserved if one reflects source class uncertainty, via the probabilities
\[{\rm P}[C_s=c] = \frac{f_{\underline{Z}|c}(\underline{z})}{\sum_{c'\neq c_d}f_{\underline{Z}|c'}(\underline{z})}\quad\forall c\neq c_d.\]
Going further, if we have knowledge of the class confusion matrix $[{\rm P}[C^*=i|C=j]]$, then a tuple $(c_s,c_d)$ with a small class confusion probability ${\rm P}[C^*=c_d|C=c_s]$ may indicate an attack. As a result, we weight KL distance by $\frac{1}{{\rm P}[c_d|c_s]}$. This increases the decision statistic for those pairs that are unlikely to occur. Combining both techniques, we construct an average weighted ADA decision statistic via
\[\sum_{c\neq c_d}{\rm P}[C=c]\frac{D_{\rm KL}(P^{(c)}||Q^{(c)})}{{\rm P}[C^*=c_d|C=c]}.\]
We can evaluate this statistic at different layers and then apply a max rule over the layers.

\paragraph{Exploiting local features:} Rather than jointly null-model all features from a layer, instead we can null-model all possible feature {\it pairs} within this layer. This accounts for possible sparsity of an attack's anomalous signature within a layer\footnote{Joint atypicality of all features in a layer may be weak if only a few features exhibit strong atypicality.}. For layer $l$ with $N$ features, there are $N_l=\binom{N}{2}$ feature pairs. For each, denoted $(Z_i,Z_j)$($i$th and $j$th features from layer $l$), we can evaluate average weighted ADA statistics. Moreover, each of these low-order AW-ADA statistics can be {\it weighted} by the magnitude of the DNN weights from $Z_i$ and $Z_j$ to the next layer of the DNN. The DNN weightings are properly normalized and denoted $\beta_i$ and $\beta_j$. The feature pairs with higher $\beta_i$ and $\beta_j$ have a stronger impact in classifier decision-making and thus atypicalities involving them should be given greater weight. Accordingly, for each layer we form a weighted aggregation of all low-order AW-ADA statistics, expressed for layer $l$ as {\it L-AWA-ADA}$^{(l)}$:
\[\frac{1}{N_l}\sum_{(i,j)}\beta_i\beta_j\sum_{c\neq c_d}{\rm P_{ij}}[C=c]\frac{D_{\rm KL}(P_{ij}^{(c)}||Q^{(c)})}{{\rm P}[C^*=c_d|C=c]}.\]
Here ${\rm P}_{ij}$ and $P_{ij}^{(c)}$ are feature-pair dependent since they are calculated using null density modelling $f(\cdot)$, which is feature-pair dependent.$\beta_i$ is the sum of the magnitudes of
the DNN weights that conduct from feature $i$ in layer
$l$ to all neurons in the next layer, $l + 1$, normalized by
the maximum such sum over all features in layer $l$. $1/N_l$ is a necessary normalizer to compare distance statistics across layers fairly, since different layers have different numbers of feautures (neurons). Again we apply a max-KL rule on {\it L-AWA-ADA}$^{(l)}$ statistics, with the resulting method dubbed {\it L-AWA-ADA-maxKL}.  This is the ultimate ADA detection method (achieving the best results), described in more detail in 
\cite{ADA-arxiv17,MLSP18-ADA}.
\section{Proposed Detection Approach for Reverse Engineering Attacks}\label{sec:ADA-RE}
Since in RE attacks the attacker submits batches of query images to the classifier, we modify L-AWA to jointly exploit batches of images in seeking to detect attacks (in this case RE query attacks, not TTEA attacks).
Several schemes for {\it aggregating} L-AWA-ADA decision statistics, produced for individual images in a batch, are investigated: 
\begin{itemize}
\item[i)] arithmetically averaging the L-AWA statistic over all images in a batch; 
\item[ii)] {\it maximizing} the L-AWA statistic over all images in a batch; 
\item[iii)] Dividing a batch into mini-batches, for example a batch of 50 images could be divided into mini-batches of size 5.  For each mini-batch, apply either scheme i) or ii).  Then, make a detection if {\it any} of the mini-batches yields a detection statistic greater than the threshold ({\it union rule}). 
\end{itemize}
This last scheme will be seen to perform the best.
\section{Experimental Setup and Results}\label{sec:expts}
\begin{figure}[h]\centering
\centering\includegraphics[width=4.35in]{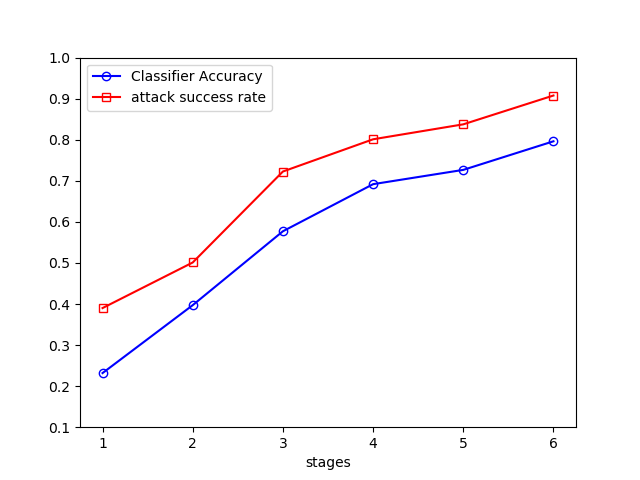}
\caption{attack sucess rate and classifier accuracy versus RE stage}\label{fig:3}
\end{figure}

We experimented on MNIST \cite{mnist}. This is a dataset with 60,000 grayscale images, representing the digits `0' through
`9'. There are 50,000 training images and 10,000 test images.  As a DNN classifier, we used Lenet-5 \cite{LeCun}. We also used the Lenet-5 structure for training the RE attacker's substitute network.
For $S_0$ we used 150 MNIST samples (15 from each class). We applied 5 stages of retraining (6 training stages) of the substitute DNN and chose $\lambda=0.1$.  The number of queries generated by the 5 stages were: 150, 300, 600, 1200 and 2400. Fast gradient sign method (FGSM) is used to craft adversarial samples.  We used mini-batches of size 5 in experiments. Two maxpooling layers and the penultimate layer were used in generating the ADA detection statistics.For evaluating RE detections ROC-AUC, we used two data sources: the 10,000 (non-query) test images and the query images produced in a given stage. For a given batch size, to crate a pool of samples used for evaluating ROC-AUC, we randomly drew batches from the two sources many times with replacement. The number of samples created was batch-size dependent. As one example for batch size 20, we created 427 non-query batches (samples) and 361 query batches (samples).
\begin{figure}[h]\centering
\centering\includegraphics[width=4.35in]{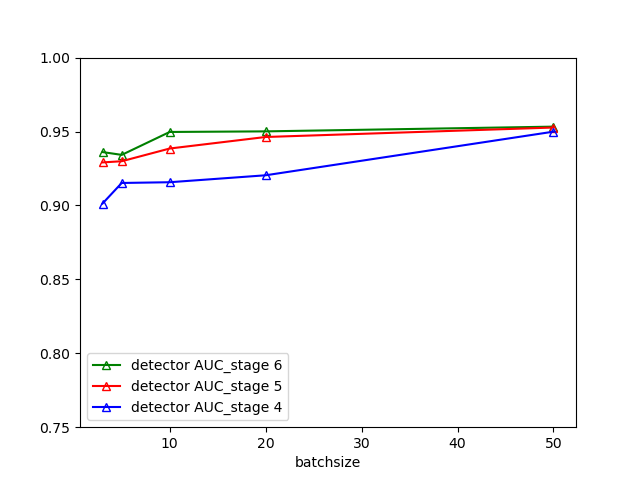}
\caption{RE detection ROC AUC at different stages versus batch size for arithmetic averaging scheme.}\label{fig:1}
\end{figure}
\begin{figure}[h]\centering
\centering\includegraphics[width=4.35in]{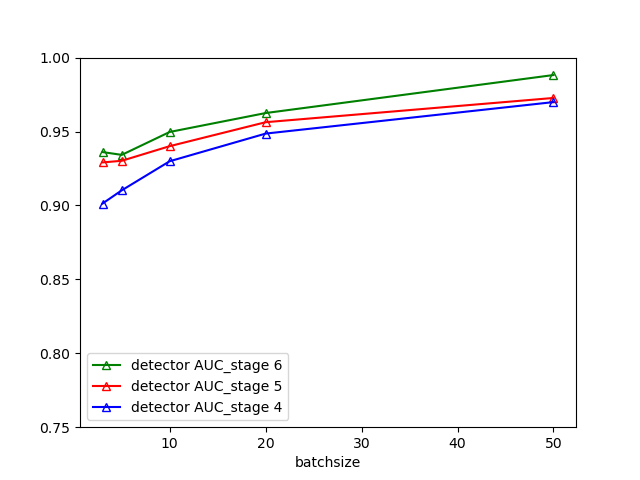}
\caption{RE detection ROC AUC at different stages versus batch size for mini-batch union aggregation scheme.}\label{fig:2}
\end{figure}
We evaluated detection accuracy for stages 4-6 in our experiments.  The reason is as follows: the substitute classifier's accuracy and the resulting success rate of TTEA attacks both grow with the stage number; by stage 4, these accuracies are 0.69 and 0.8, respectively, as shown in Figure \ref{fig:3}.
Figure \ref{fig:1} shows that good detection accuracy is achieved using the arithmetic averaging scheme, with the ROC AUC increasing with batch size
and with the attack stage, as expected (slightly inferior performance is achieved by max rule).  However, the ROC AUC appears to asymptote at about 0.95, which we would not expect -- we would hope perfect detection accuracy could be approached with increasing batch size, especially in the latter stages.  This better
behaviour is exhibited by the mini-batch scheme with union detection rule in Figure \ref{fig:2}.  Thus, this latter aggregation scheme
is the most promising one.
\section{Conclusion}
We have developed an anomaly detection scheme that is very effective at detecting ``stealthy" RE attacks on DNN image classifiers. This is potentially important to protect black box classifier information and to prevent TTEAs.
Detection of other types of attacks, for other application domains, may be considered in future.

\bibliographystyle{plain}
\bibliography{../../latex/adversarial,../../latex/adversarial0,../../latex/kesidis-prior}
\end{document}